\documentclass[runningheads]{llncs}

\usepackage{algorithm}
\usepackage{algorithmic}
\usepackage{amsmath}
\usepackage{booktabs}
\usepackage{multirow} 
\usepackage{hyperref}

\usepackage[T1]{fontenc}

\usepackage{graphicx,verbatim}

\newcommand{\net}{\mathbf{f}_\theta}

\begin{document}

\title{CAPE: Connectivity-Aware Path Enforcement Loss for Curvilinear Structure Delineation}

\titlerunning{CAPE: Connectivity-Aware Path Enforcement}

\author{Elyar Esmaeilzadeh\orcidID{0009-0003-5610-2604}\and
Ehsan Garaaghaji\orcidID{0009-0003-2720-7415}\and
Farzad Hallaji Azad\orcidID{0009-0003-3882-4275}\and
Doruk Oner \thanks{Corresponding author: doruk.oner@bilkent.edu.tr}\orcidID{0000-0002-9403-4628}}

\authorrunning{Esmaeilzadeh et al.}
\institute{NeuraVision Lab, Department of Computer Engneering, \\ Bilkent University, Ankara, Turkey \\
    \email{\{elyar,ehsan.garaaghaji,farzad.hallaji,doruk.oner\}@bilkent.edu.tr}}

\maketitle              


\begin{abstract}
Promoting the connectivity of curvilinear structures, such as neuronal processes in biomedical scans and blood vessels in CT images, remains a key challenge in semantic segmentation. Traditional pixel-wise loss functions, including cross-entropy and Dice losses, often fail to capture high-level topological connectivity, resulting in topological mistakes in graphs obtained from prediction maps. In this paper, we propose CAPE (Connectivity-Aware Path Enforcement), a novel loss function designed to enforce connectivity in graphs obtained from segmentation maps by optimizing a graph connectivity metric. CAPE uses the graph representation of the ground truth to select node pairs and determine their corresponding paths within the predicted segmentation through a shortest-path algorithm. Using this, we penalize both disconnections and false positive connections, effectively promoting the model to preserve topological correctness. Experiments on 2D and 3D datasets, including neuron and blood vessel tracing demonstrate that CAPE significantly improves topology-aware metrics and outperforms state-of-the-art methods. Code is available at \href{https://github.com/NeuraVisionLab/CAPE}{https://github.com/NeuraVisionLab/CAPE}.


\end{abstract}

\section{Introduction}

Segmenting curvilinear structures—such as neuronal processes in biomedical scans and blood vessels in CT images—is essential for numerous applications in medical imaging and biological research. Traditional deep convolutional networks have significantly improved pixel-level segmentation. However, despite these advancements, they exhibit a fundamental limitation: they are trained using pixel-wise loss functions such as cross-entropy, Dice, or Mean Squared Error (MSE), which do not explicitly enforce topological correctness, because they are designed to optimize the number of correctly segmented pixels/voxels, but do not account for high-level structural continuity. This is particularly problematic, as connectivity errors can severely compromise the reliability of constructions and their usability in downstream tasks. As a result, annotators are often required to manually correct topological errors, a highly labor-intensive and tedious process, especially when dealing with large volumetric datasets.

Recent approaches have focused on designing loss functions specifically aimed at enforcing topological connectivity \cite{Mosinska2017,Edelsbrunner2008,Hu2019,Clough2022,Byrne2020,Shit2020,Oner2020,Oner2022}. Some methods analyze the structural topology using Persistent Homology \cite{Edelsbrunner2008,Hu2019,Clough2022,Byrne2020}, while others incorporate perceptual loss by leveraging pre-trained feature extractors \cite{Mosinska2017}. Additionally, certain approaches promote connectivity by enforcing background separation through Rand index optimization \cite{Oner2020,Oner2022}. Another strategy involves developing differentiable loss functions based on existing metrics \cite{Shit2020}. Despite these efforts, all existing methods attempt to achieve global connectivity by optimizing a metric that is not explicitly designed for connectivity. Consequently, they only indirectly enforce the global topological correctness of predictions.

In this paper, we propose CAPE (Connectivity-Aware Path Enforcement), a novel loss function that improves the connectivity of curvilinear structure reconstructions. Inspired by the Average Path Length Similarity (APLS) metric \cite{etten2017spacenet}, CAPE evaluates connectivity by comparing entire paths instead of individual pixels. Specifically, it leverages Dijkstra’s algorithm to compute shortest paths in the pixel domain by mapping paths between nodes from the ground truth graph onto the predicted segmentation. This direct comparison enforces high-level topological correctness, leading to improved segmentation.


\section{Related Work}

Over the years, various approaches have been proposed for delineating curvilinear structures, ranging from hand-designed filters \cite{Frangi,Law,Turetken2013} to learned filters \cite{Wu,Breitenreicher} using support vector machines \cite{Huang20042009}, gradient boosting \cite{Sironi_2014_CVPR}, and decision trees \cite{Turetken_Engin}. However, these traditional methods suffer from limited generalizability and various shortcomings. With the advancements in deep learning, more powerful deep neural networks have largely replaced these techniques \cite{Mnih,Ganin,Maninis,Peng,Guo,WOLTERINK201946}.

Recent advances in biomedical image segmentation have been driven by deep convolutional neural networks, with U-Net \cite{Ronneberger2015} being especially influential. While U-Net achieves high pixel-level accuracy, conventional pixel-wise loss functions neglect topological correctness—a critical issue for curvilinear structures like neuronal processes, where connectivity errors degrade segmentation quality. Consequently, many studies have explored methods beyond pixel-wise losses to fully harness the potential of deep networks.

Several studies have designed novel topology-aware loss functions to improve segmentation performance of deep networks. \cite{Mosinska2017} penalized differences in label and prediction feature maps using a pretrained VGG19 network \cite{simonyan2014very}, but it lacks explicit topological preservation. Persistent Homology-based methods \cite{Edelsbrunner2008,Hu2019,Clough2022} provide differentiable topological similarity but ignore spatial alignment, leading to inaccuracies. \cite{Oner_Persistent} introduced a filtration function addressing this issue, though persistent homology remains computationally expensive. The soft-clDice loss \cite{Shit2020} enhances connectivity via soft-skeletonization but remains pixel-wise and volumetric-focused. MALIS loss \cite{Funke2017} learns region affinities but struggles with curvilinear structures, because of the existance of loops in those structures. \cite{Oner2020} proposed inverse MALIS for curvilinear connectivity by enforcing regional separation, effective in 2D but not generalizing to 3D. To address this, \cite{Oner2022} introduced a projection-based extension for 3D, yet it relies on Rand Index, which is suboptimal for curvilinear structures and yields sparse gradients, limiting global connectivity enforcement. Additionally, it often penalizes a single pixel per background region, resulting in highly sparse gradients, making the training less stable.

To evaluate networks for curvilinear structure segmentation, we use metrics beyond the Rand Index, such as \textit{Average Path Length Similarity} (APLS) \cite{etten2017spacenet}. APLS, a graph-based metric, quantifies similarity by comparing shortest path distances between corresponding nodes, penalizing deviations in length and structure to assess connectivity. Unlike pixel-wise metrics, connectivity-based metrics evaluate groups of pixels or voxels, making conventional metrics ineffective for topological correctness. However, using APLS and TLTS \cite{TLTS} as loss functions is impractical due to their non-differentiability in the graph domain.

While existing methods achieve competitive results, none directly optimize connectivity-based metrics. CAPE, inspired by APLS, introduces a differentiable, pixel-domain variant. Unlike prior methods \cite{Oner2020,Oner2022}, which yield sparse gradients penalizing individual pixels, CAPE produces denser gradients along the entire path, enhancing connectivity enforcement while remaining suitable for gradient-based optimization.


\section{Methodology}

Our approach is inspired by the APLS connectivity metric, which evaluates the connectivity of curvilinear structures but is inherently non-differentiable and unsuitable for direct integration into training segmentation networks. In APLS, edges are binary—either present or absent. In our differentiable adaptation, we replace these binary affinities with continuous ones (distance values predicted by our network). This allows path lengths to be computed as sums of network predictions, enabling optimization. At a high level, our loss function measures the consistency of connectivity between the ground truth and the network’s prediction by comparing the costs of corresponding shortest paths. Specifically, we consider the ground truth graph \(G=(V,E)\), where vertices \(V\) represent key structural points and edges \(E\) capture the true connections among foreground elements. Our network, \(\net{}\), predicts a distance map \(\hat{\mathbf{y}}\) in which each pixel indicates its distance from the nearest foreground structure (for brevity, we refer to image elements as pixels; in 3D data, these correspond to voxels, and the same methodology applies). We predict distance maps instead of probability maps because distance map pixel values rise linearly as disconnections increase, more effectively penalizing longer or severe gaps. Probability maps, however, saturate near zero in disconnected regions, reducing loss sensitivity.

\subsection{CAPE Loss}

In an iterative procedure, we randomly select two connected vertices \(v_1\) and \(v_2\) from \(G\) and compute their ground truth shortest path, \(\text{path}_G\), using Dijkstra’s algorithm. We then project these vertices onto the predicted distance map by selecting the local minima within a constrained 7×7 neighborhood to account for small deviations between \(G\) and the prediction centerlines, yielding refined positions \(v'_1\) and \(v'_2\). A second Dijkstra pass on the predicted distance map computes the predicted shortest path, \(\text{path}_{\hat{\mathbf{y}}}\). The cost of \(\text{path}_{\hat{\mathbf{y}}}\) is defined as,
\begin{equation}
\text{cost}(\text{path}_{\hat{\mathbf{y}}}) = \sum_{n \in \text{path}_{\hat{\mathbf{y}}}} \hat{\mathbf{y}}(n)^2,
\end{equation}
where $n$ is the set of pixels in \(\text{path}_{\hat{\mathbf{y}}}\), and \(\mathbf{\hat{y}}(n)\) is the predicted distance value for pixel \(n\). We use the square of the distance predictions to ensure that pixels with higher distance values—indicating more severe disconnections—contribute more significantly to the overall loss, thereby generating stronger gradients compared to those with mild disconnections. This cost is fully differentiable with respect to the network parameters, making our loss function well-suited for gradient-based optimization. 

\begin{figure}[h]
\centering
\includegraphics[width=0.89\textwidth]{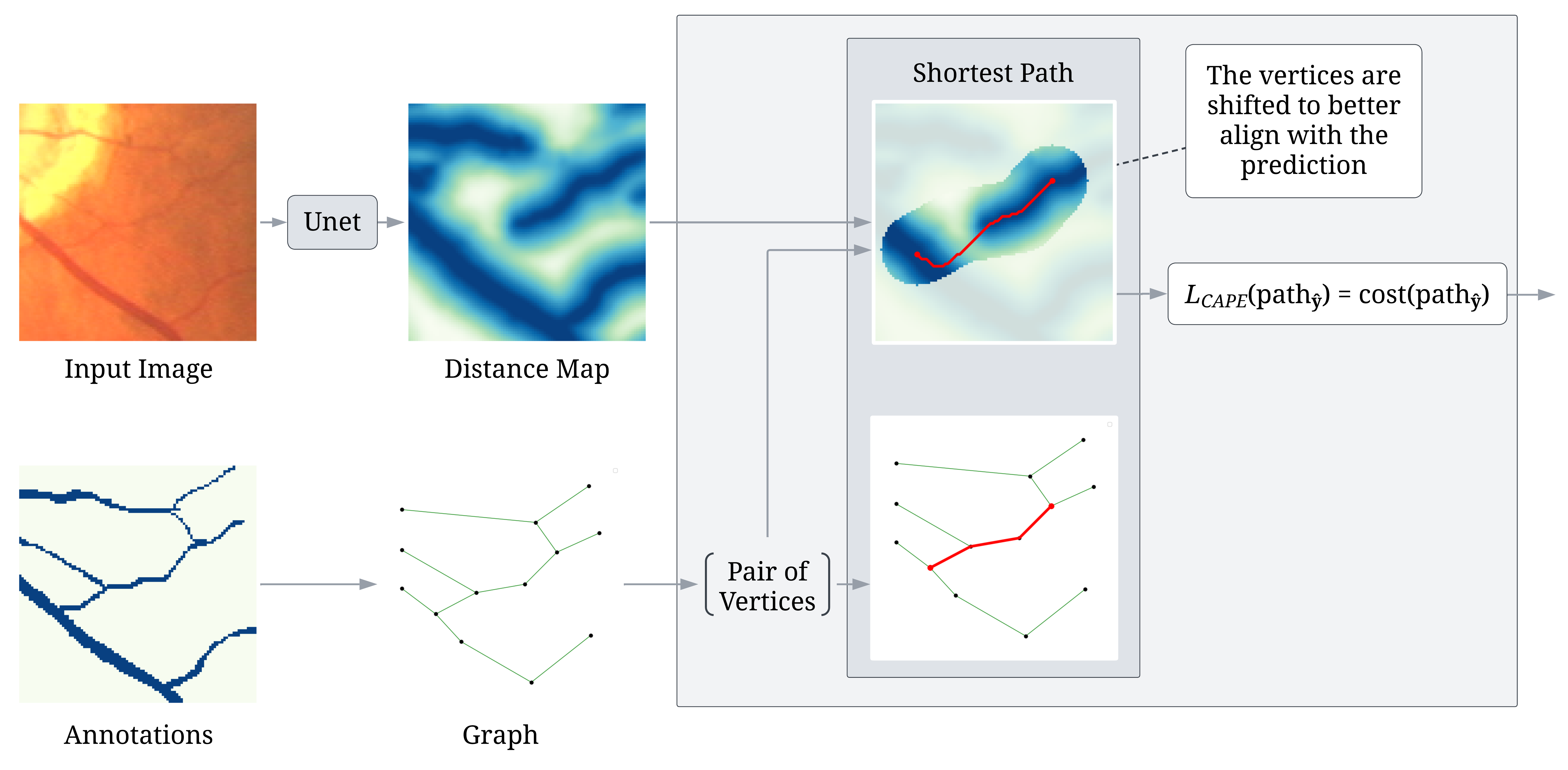}
\caption{\textbf{Computation of $L_{CAPE}$.} After extracting the ground truth graph, an iterative process selects pairs of vertices and computes their shortest path. The corresponding path is then masked with dilation and then projected to the pixel domain, and the shortest path algorithm is reapplied to obtain $L_{CAPE}$.} \label{method_fig}
\end{figure}

For a corresponding pair of paths from the ground truth, \(\text{path}_G\), and prediction, \(\text{path}_{\hat{\mathbf{y}}}\), the loss is defined as the absolute difference between their costs. Because the ground truth graph embodies perfect connectivity, we define the cost of any ground truth path \(\text{path}_G\) as 0, which simplifies the loss to cost of \(\text{path}_{\hat{\mathbf{y}}}\)
\begin{equation}
L_{\textsc{CAPE}}(\text{path}_G, \text{path}_{\hat{\mathbf{y}}}) = \left| \text{cost}(\text{path}_G) - \text{cost}(\text{path}_{\hat{\mathbf{y}}}) \right| = \text{cost}(\text{path}_{\hat{\mathbf{y}}}).
\end{equation}
If the network confidently captures the connection, the cost along the predicted path will be close to zero, yielding a low loss; conversely, if there is a disconnection, high distance values along the path will increase the cost and result in a higher loss.

The iterative procedure continues until all connections in \(G\) are evaluated. After processing each pair, the corresponding edges are removed from \(G\) to ensure unique processing, and the procedure repeats until no edges remain. The complete procedure for computing the loss can be found in Algorithm~\ref{alg:cape_algo}.


\begin{algorithm}
\caption{Computation of \(L_{\textsc{CAPE}}\)}\label{alg:cape_algo}
\begin{algorithmic}[1]
\renewcommand{\algorithmicrequire}{\textbf{Input:}}
\renewcommand{\algorithmicensure}{\textbf{Output:}}

\REQUIRE Groundtruth Graph \( G = (V, E) \) and predicted distance map \( \hat{\mathbf{y}} \)

\ENSURE \( L_{\textsc{CAPE}}(G, \hat{\mathbf{y}}) \)
\STATE Initialize \( L_{\textsc{CAPE}} \gets 0 \)
    \WHILE{\( E \neq \emptyset \)}
        \STATE Pick a pair of vertices, \( (v_1, v_2) \), that are connected from \(G\)
        \STATE \( path_{G} \gets \text{Dijkstra}(G, v_1, v_2) \)
        \STATE \(v'_1, v'_2 \gets \text{shift}(v_1), \text{ shift}(v_2) \)
        \STATE \( M \gets \text{render}(\text{path}_{G})\) 
        \STATE \(M_{\text{dilated}} \gets \text{dilate}(M)\)
        \STATE \( path_{\hat{\mathbf{y}}} \gets \text{Dijkstra}(\hat{\mathbf{y}} \cdot M_{\text{dilated}}, v'_1, v'_2) \)
        \STATE \( L_{\textsc{CAPE}} \gets L_{\textsc{CAPE}} + cost(\text{path}_{\hat{\mathbf{y}}})\)
        \FOR{\( e \in \text{path}_{G} \)}
            \STATE Remove \( e \) from \( E \)
        \ENDFOR
    \ENDWHILE
\RETURN \( L_{\textsc{CAPE}} \)
\end{algorithmic}
\end{algorithm}

The reason we perform a second Dijkstra on the predicted distance map rather than directly using the ground truth shortest path for cost computation is to allow for minor spatial deviations in the predicted path. 
Centerline annotations can be noisy, and enforcing exact spatial correspondence would undesirably penalize predictions that correctly capture connectivity despite small shifts. 
By computing the predicted path cost via a separate Dijkstra pass on \(\hat{\mathbf{y}}\), our loss function focuses on evaluating connectivity rather than strict spatial alignment. The computation of loss is explained visually in Figure~\ref{method_fig}.

\begin{figure}[h]
\centering
\includegraphics[width=0.9\textwidth]{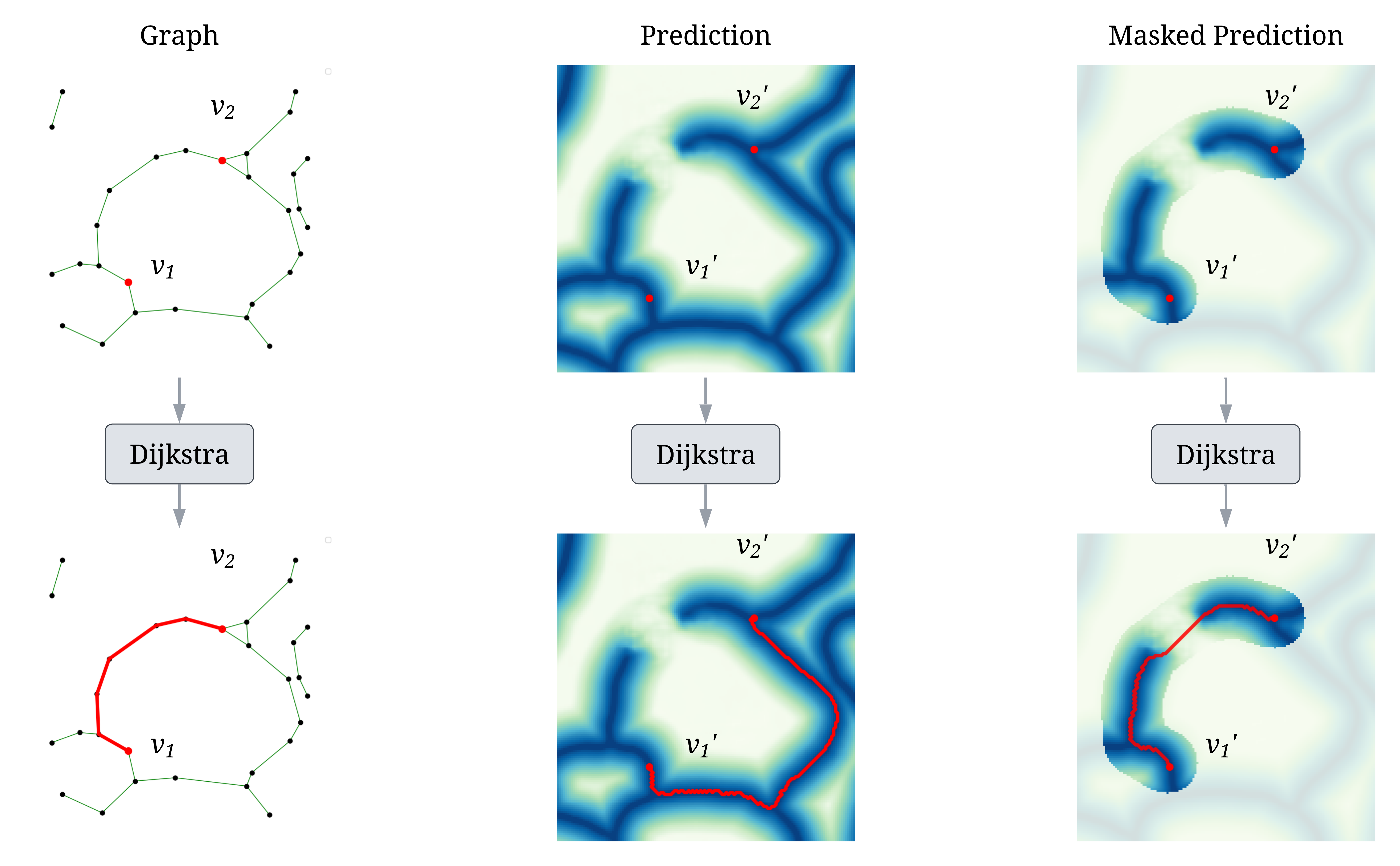}
\caption{\textbf{Masking Strategy.} Left: Ground truth graph with selected vertices and the shortest path computed via Dijkstra’s algorithm. Middle: Projected vertices on the predicted distance map and the corresponding Dijkstra path computed without masking; due to a loop, the algorithm bypasses the disconnection. Right: With a mask applied around the ground truth path, Dijkstra is forced to capture the disconnection by preventing an alternate loop path.} \label{masking_fig}
\end{figure}

In addition, we incorporate a masking strategy to further constrain the search for the predicted path. First, we render the ground truth path to obtain a binary mask \(M\), then dilate this mask by 10 pixels to create a dilated mask \(M_{\text{dilated}}\).

The dilated mask is applied element-wise to the predicted distance map \(\hat{\mathbf{y}}\), and the shortest path is computed within the masked region between the projected vertices \(v'_1\) and \(v'_2\) using Dijkstra’s algorithm:
\begin{equation}
\text{path}_{\hat{\mathbf{y}}} = \text{Dijkstra}(\hat{\mathbf{y}} \cdot M_{\text{dilated}}, v'_1, v'_2).
\end{equation}
This approach confines the search to a region closely aligned with the ground truth, allowing only limited deviations and reducing the likelihood of selecting an alternative, longer path—especially in cases where loops are present—while also lowering the computational complexity of the loss calculation. As illustrated in Figure~\ref{masking_fig}, masking effectively guides the selection of matching paths, ensuring that the intended shortest path is chosen even in loopy structures.

\subsection{Training Loss Integration}

The total CAPE loss, \(L_\textsc{CAPE}(G,\hat{\mathbf{y}})\), is defined as the sum of the losses computed for each ground truth path sampled from the graph \(G\). 
\begin{equation}
L_{\textsc{CAPE}}(G, \hat{\mathbf{y}}) = \sum_{\text{path}_G \in \mathcal{P}_G} L_{\textsc{CAPE}}(\text{path}_G, \text{path}_{\hat{\mathbf{y}}}) = \sum_{\text{path}_G \in \mathcal{P}_G} \text{cost}(\text{path}_{\hat{\mathbf{y}}}),
\end{equation}
where \(\mathcal{P}_G\) is the set of paths sampled from \(G\).

In our training framework, we combine the CAPE loss with a per-pixel Mean Squared Error (MSE) loss to provide comprehensive supervision. While the CAPE loss delivers connectivity-aware gradients primarily along foreground paths, the MSE loss ensures dense supervision across all pixels, including background regions. The total loss function is defined as
\begin{equation}
L_{\textsc{total}} = L_{\text{MSE}}(\mathbf{y}, \hat{\mathbf{y}}) + \alpha L_{\text{CAPE}}(G, \hat{\mathbf{y}}),
\end{equation}
where \(\mathbf{y}\) is the ground truth per-pixel distance map computed from the graph \(G\) and \(\alpha\) is a hyperparameter to balance the per-pixel and connectivity-aware components of the loss.



\begin{figure*}[hbt!]
\centering
\setlength{\tabcolsep}{3pt}
\begin{tabular}{c c c c c c c}
	& \scriptsize{\textbf{Input}} & \scriptsize{\textbf{Label}} & \scriptsize{\textbf{Perc}} & \scriptsize{\textbf{clDice}} & \scriptsize{\textbf{InvMALIS}} & \scriptsize{\textbf{CAPE}}\\
    
	\raisebox{6.5mm}{\rotatebox[origin=t]{90}{\scriptsize{\textbf{CREMI}}}} &
	\includegraphics[width=0.14\textwidth]
    {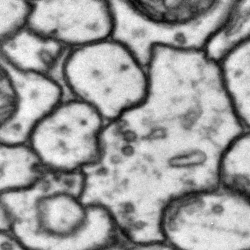} &
	\includegraphics[width=0.14\textwidth]
    {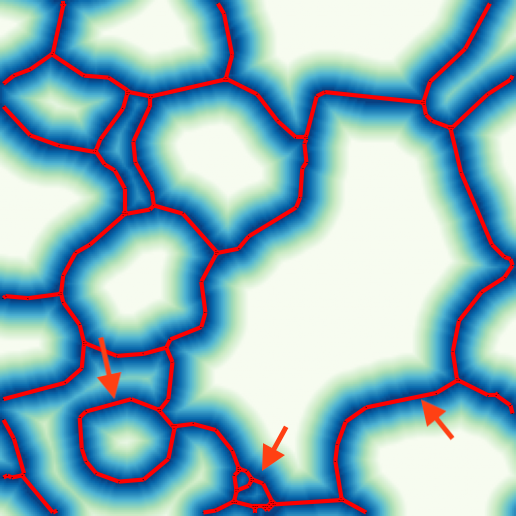} &
	\includegraphics[width=0.14\textwidth]
    {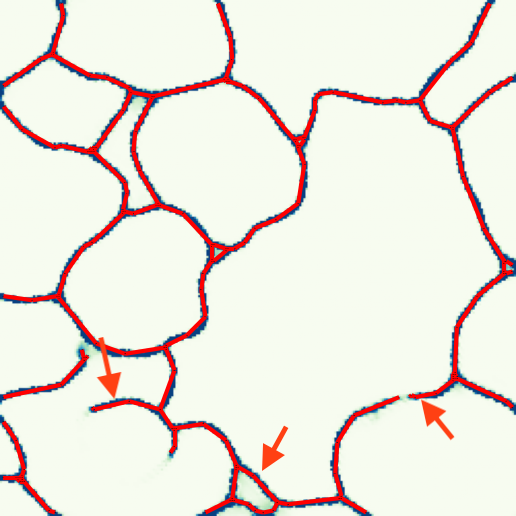} &
    \includegraphics[width=0.14\textwidth]
    {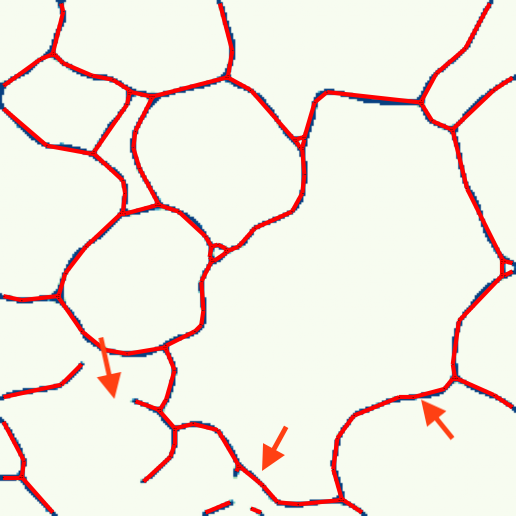} &
    \includegraphics[width=0.14\textwidth]
    {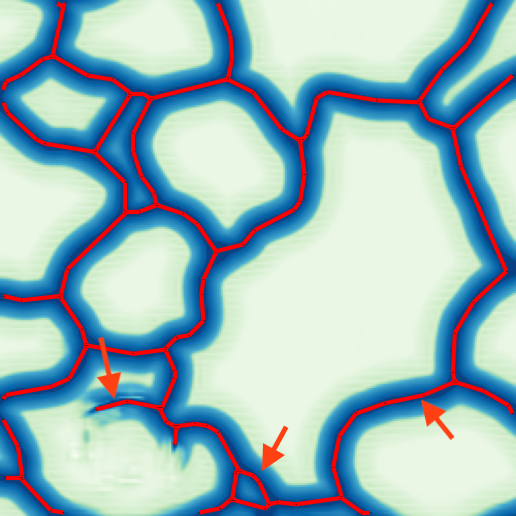} &
	\includegraphics[width=0.14\textwidth]
    {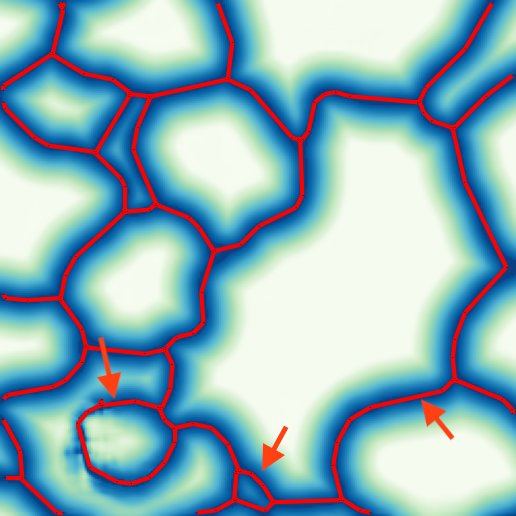}\\

	\raisebox{6.5mm}{\rotatebox[origin=t]{90}{\scriptsize{\textbf{DRIVE}}}} &
	\includegraphics[width=0.14\textwidth]
    {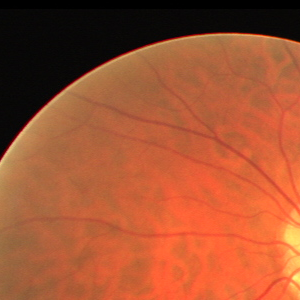} &
	\includegraphics[width=0.14\textwidth]
    {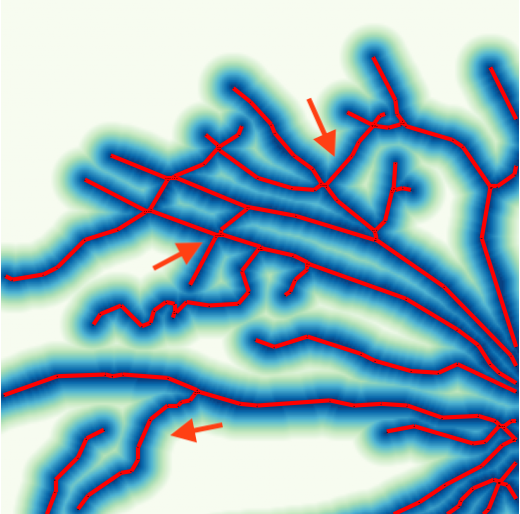} &
	\includegraphics[width=0.14\textwidth]
    {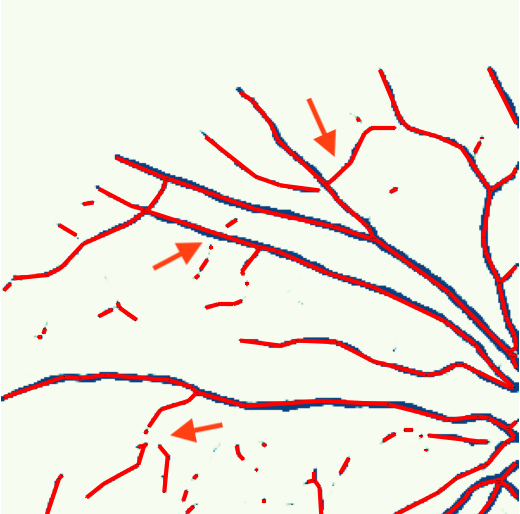} &
    \includegraphics[width=0.14\textwidth]
    {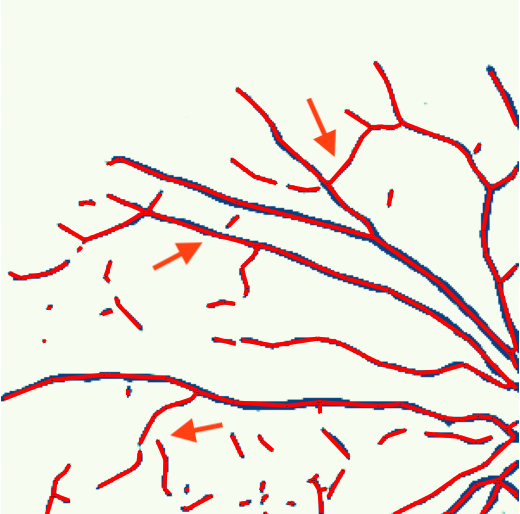} &
    \includegraphics[width=0.14\textwidth]
    {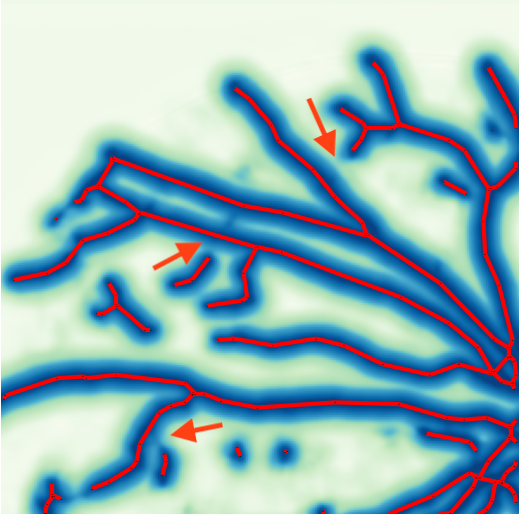} &
	\includegraphics[width=0.14\textwidth]
    {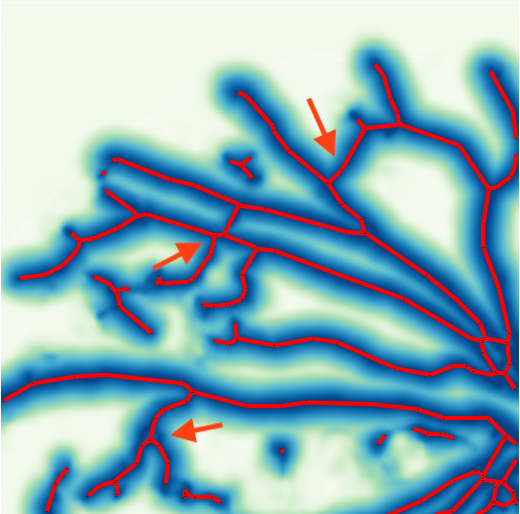} \\

    \raisebox{6.5mm}{\rotatebox[origin=t]{90}{\scriptsize{\textbf{Brain}}}} &
	\includegraphics[width=0.14\textwidth]
    {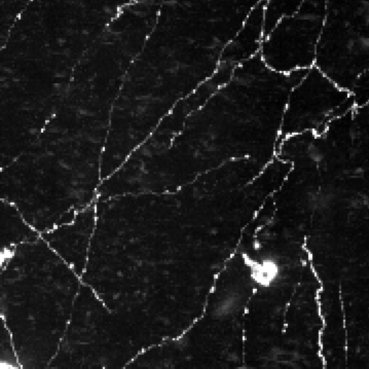}&
	\includegraphics[width=0.14\textwidth]
    {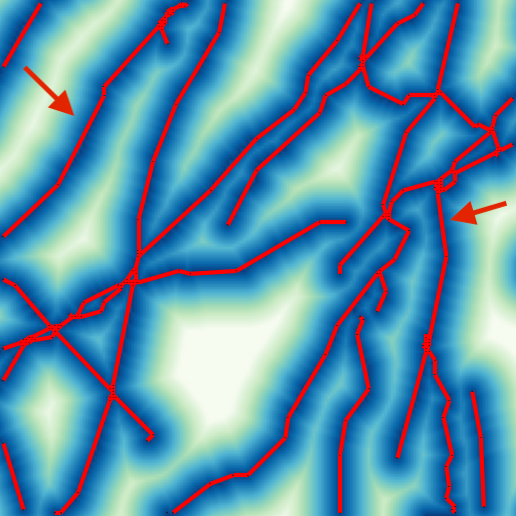} &
	\includegraphics[width=0.14\textwidth]
    {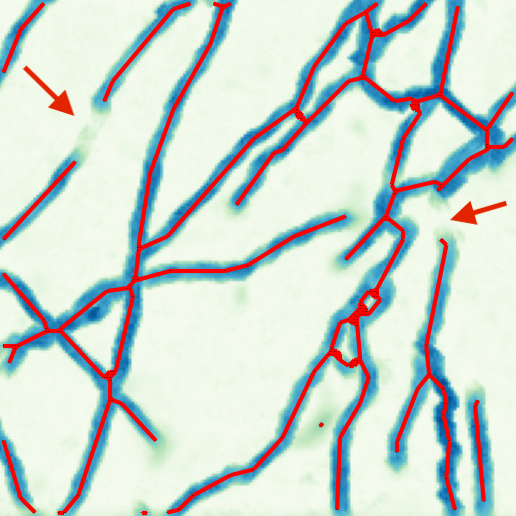} &
    \includegraphics[width=0.14\textwidth]
    {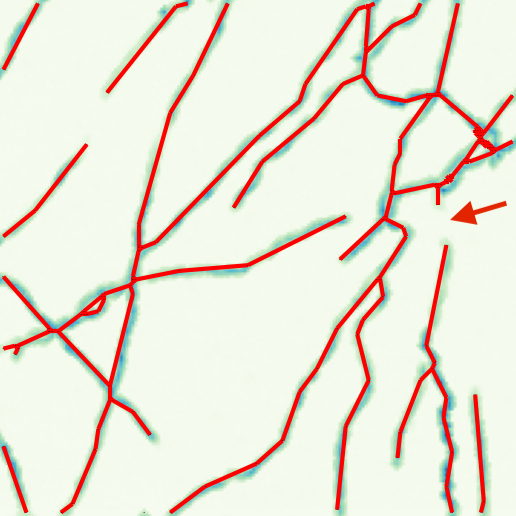} &
    \includegraphics[width=0.14\textwidth]
    {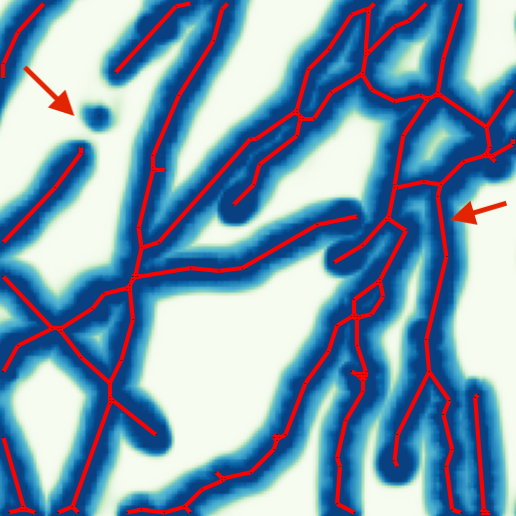} &
	\includegraphics[width=0.14\textwidth]
    {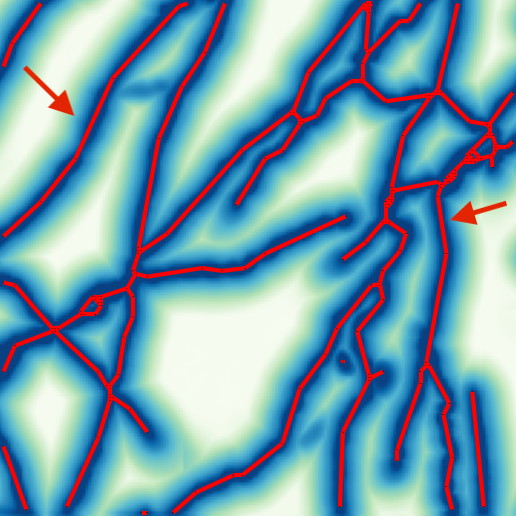} \\

\end{tabular}

\caption{
Qualitative comparison of the test results in 2D and 3D datasets. The arrows highlight regions that remain disconnected in alternative methods but are successfully connected using CAPE. The connectivity improves significantly when our approach is used.
\label{fig_results}
}

\end{figure*}

\section{Experiments}

\begin{table}[h]
\centering
\caption{
Comparative results.  
As seen in the table, our method not only gives state-of-the-art results in the topology-aware metrics, but also comparative results in pixel-wise metrics.
\label{tab:results}
}

\setlength{\tabcolsep}{5pt}

\begin{tabular}{@{} l l c c   c  c c  c  c@{} }
\toprule
 && \multicolumn{4}{c}{Pixel-wise} & &  \multicolumn{2}{c}{Topology-aware}\\
 \cmidrule{3-6}
 \cmidrule{8-9}
Dataset  &  Methods  &     
Corr. &      Comp. &       Qual.  & Dice &  &    APLS &        TLTS \\
\midrule
\multirow{5}{*}{{CREMI}}
						&MSE &
        							\underline{98.3} &        99.6 &        97.9 & 85.4 &  &     81.9 &       79.5 \\ 
						&Perc &
        							{\bf 99.9} &        98.8 &        {\bf 98.7} & {\bf 88.9} & &    91.6 &        91.2 \\
						&clDice &
        							99.0 &        96.5 &      95.7     & 79.3 &   &     82.3 &        81.2 \\
        				&InvMALIS  &
        							98.2&        \underline{99.7} &        97.9 & 87.9 &  &   \underline{91.9} &        \underline{92.4} \\
                            &OURS  &
        							\underline{98.3} &        {\bf 99.8} &        \underline{98.1} & \underline{88.7} &    &    {\bf 93.6} &        {\bf 93.5} \\
\midrule
\multirow{5}{*}{{DRIVE}}
						&MSE &
        							{\bf 97.2} &        92.6 &        \underline{90.4} & 77.1 &  &     68.2 &       66.9 \\ 
						&Perc &
        							\underline{96.3} &        93.5 &        \underline{90.4} & 77.3 & &      76.8 &        72.8 \\
						&clDice &
        							94.2 &        93.8 &      88.8     & 76.2 &  &     74.5 &        70.5 \\
        				&InvMALIS  &
        							94.6&        {\bf 94.7} &        89.9 &  \underline{79.3} &  &    \underline{77.7} &        \underline{74.5} \\
                            &OURS  &
        							95.2&        \underline{94.6} &        {\bf 90.5} & {\bf 80.3} &   &    {\bf 80.2} &        {\bf 76.4} \\
\midrule
\multirow{5}{*}{{Brain}}
						&MSE &
        							\bf{99.0} &        94.1 &        93.2 & 75.2 & &     83.9 &       75.2 \\ 
						&Perc &
        							97.1 &        \bf{97.3} &        \underline{94.6} & \bf{80.4} & &      83.5 &        \underline{86.4} \\
						&clDice &
        							\underline{98.3} &        96.5 &      \bf{94.9}     & 70.6 &  &     84.4 &        85.3 \\
        				&InvMALIS  &
        							96.6&        96.2 &        93.0 & 68.2 &   &    \underline{85.7} &        83.8 \\
                            &OURS  &
        							97.1&        \underline{97.2} &        94.4 &  \underline{78.3} &  &    \bf{89.8} &        \bf{93.1} \\
\bottomrule
\end{tabular}

\end{table}
\subsection{Datasets}

We conduct experiments on 2D and 3D datasets. For 2D experiments, we use two datasets: \textbf{CREMI} \cite{Funke2017}, which contains 83 training and 42 validation samples of size $1250\times 1250$, including neurons from the \textit{Drosophila melanogaster} brain captured by serial section electron microscopy, and the \textbf{DRIVE} dataset \cite{DRIVE}, which contains 13 training and 7 validation samples from digital retinal images of blood vessels, sized $584\times 565$. For the 3D experiments, we use the \textbf{Brain} dataset, which consists of 14 light microscopy scans of the mouse brain, each of size $250\times 250\times 200$. We use 10 of them for training and 4 for validation.

\subsection{Metrics}

We evaluate our method using two pixel-wise metrics, \textbf{CCQ} and \textbf{Dice}, and two connectivity-aware metrics, \textbf{APLS} and \textbf{TLTS}. \textbf{CCQ} (Correctness, Completeness, and Quality) is analogous to precision, recall, and F1-score, with true positives defined as pixels within 3 units of the ground truth, making it well-suited for centerline segmentation. \textbf{Dice} score measures the overlap between predicted and true segmentations. For connectivity, \textbf{APLS} (Average Path Length Similarity) compares the average shortest path lengths between corresponding nodes in the ground truth and predicted graphs, while \textbf{TLTS} (Too-Long-Too-Short) quantifies the fraction of predicted paths whose lengths deviate by less than 15\% from the ground truth.

\subsection{Architectures and Baselines}

For our 2D experiments, we use a 2D U-Net~\cite{Ronneberger2015}, featuring three down-samplings, two convolutional layers per level, max-pooling for the encoder, and bilinear up-sampling for the decoder. For 3D experiments, we employ a 3D U-Net with the same configuration. 
The 2D models are trained for 10k epochs and the 3D models are trained for 50k epochs, both with Adam optimizer~\cite{ADAM} with a learning rate of $1e-3$ and weight decay of $1e-3$.
We compare our results against \textbf{MSE} loss ($L_{MSE}$) between predictions and ground truths, a network trained with \textbf{Perceptual loss} using a pre-trained VGG19~\cite{Mosinska2017,simonyan2014very}, \textbf{clDice} loss~\cite{Shit2020}, and $\textbf{InvMALIS}$. For 2D datasets, we compare to inverse MALIS proposed in \cite{Oner2020}, and for 3D, we compare to the inverse MALIS computed on projections \cite{Oner2022}.

\subsection{Results}

Figure~\ref{fig_results} and Table~\ref{tab:results} present our qualitative and quantitative results, respectively. As shown in Table 1, our method significantly improves topology-aware metrics and outperforms state-of-the-art approaches, while achieving comparable performance on pixel-wise metrics. The qualitative results further highlight the enhanced connectivity in our segmentations.


\section{Conclusion and Future Work}

In this paper, we introduce CAPE (Connectivity-Aware Path Enforcement), a loss function that optimizes topological correctness by comparing ground truth and predicted shortest paths via Dijkstra’s algorithm. By generating denser gradients along entire paths, CAPE effectively penalizes disconnections and promotes connectivity. Experiments on 2D and 3D datasets demonstrate significant improvements in connectivity metrics, confirming its effectiveness for curvilinear structure segmentation in biomedical imaging. In future work, we plan to have the network directly output graph representations rather than distance maps, enabling CAPE to operate entirely in the graph domain. This unified approach is expected to further enhance topological accuracy and connectivity enforcement.

\subsubsection{\discintname}
The authors have no competing interests to declare that are
relevant to the content of this article.

\bibliographystyle{splncs04}
\bibliography{Paper-4309}  

\end{document}